%% file: main-ral.tex
\documentclass[letterpaper, 10 pt, conference]{IEEEtran}
\IEEEoverridecommandlockouts  

\usepackage{graphicx}
\usepackage{amsmath}
\usepackage{amssymb}
\usepackage{booktabs}
\usepackage{subfiles}
\usepackage{multirow}
\usepackage{soul}
\usepackage[dvipsnames]{xcolor}
\usepackage{multicol}
\usepackage{colortbl}
\usepackage{caption}
\usepackage{subcaption}
\usepackage{tikz}
\usepackage{anyfontsize}
\usepackage{amsfonts}

\graphicspath{ {./img/} }

\usepackage{algorithm}
\usepackage{algorithmic}
\usepackage{flushend}
\usepackage{newfloat}
\usepackage{listings}
\usepackage[textsize=footnotesize,colorinlistoftodos,prependcaption]{todonotes}

\input{commands}

\definecolor{cvprblue}{rgb}{0.21,0.49,0.74}
\definecolor{bluish}{rgb}{0.19, 0.55, 0.91}
\usepackage[pagebackref,breaklinks,colorlinks,citecolor=blue]{hyperref}

\usepackage[pagebackref,breaklinks,colorlinks]{hyperref}

\usepackage[capitalize]{cleveref}
\crefname{section}{Sec.}{Secs.}
\Crefname{section}{Section}{Sections}
\Crefname{table}{Table}{Tables}
\crefname{table}{Tab.}{Tabs.}

\title{Regularizing Self-supervised 3D Scene Flows\\with Surface Awareness and Cyclic Consistency}

\author{Patrik Vacek$^1$, David Hurych$^2$, Karel Zimmermann$^1$, Patrick P{\'e}rez$^3$, and Tom{\'a}{\v s} Svoboda$^1$%
\thanks{The research leading to these results has received funding from the Czech Science Foundation under Project GA 24-12360S.
This work was co-funded by the EU under the project
Robotics and advanced industrial production (reg. no. CZ.02.01.01/00/22\_008/0004590).
This research received the support of EXA4MIND project, funded by a European Union´s Horizon Europe Research and Innovation Programme, under Grant Agreement N° 101092944. Views and opinions expressed are however those of the author(s) only and do not necessarily reflect those of the European Union or the European Commission. Neither the European Union nor the granting authority can be held responsible for them. 
P.Vacek was also supported by 
Grant Agency of the CTU in Prague under Project SGS24/096/OHK3/2T/13. Valeo support is greatly acknowledged.}
\thanks{$^{1}$The authors are with the Department of Cybernetics, Faculty of Electrical Engineering, Czech Technical University in Prague, P. Vacek is the corresponding author: vacekpa2@fel.cvut.cz}
\thanks{$^2$David Hurych is with Valeo.ai}
\thanks{$^3$Patrick P{\'e}rez contributed while he was with Valeo.ai, now at Kyutai}
}

\begin{document}

\maketitle


\begin{abstract}
Learning without supervision how to predict 3D scene flows from point clouds is essential to many perception systems. We propose a novel learning framework for this task which improves the necessary regularization. Relying on the assumption that scene elements are mostly rigid, current smoothness losses are built on the definition of ``rigid clusters" in the input point clouds. The definition of these clusters is challenging and has a significant impact on the quality of predicted flows. We introduce two new consistency losses that enlarge clusters while preventing them from spreading over distinct objects. In particular, we enforce \emph{temporal} consistency with a forward-backward cyclic loss and \emph{spatial} consistency by considering surface orientation similarity in addition to spatial proximity. The proposed losses are model-independent and can thus be used in a plug-and-play fashion to significantly improve the performance of existing models, as demonstrated on two most widely used architectures. We also showcase the effectiveness and generalization capability of our framework on four standard sensor-unique driving datasets, achieving state-of-the-art performance in 3D scene flow estimation. Our codes are available on \url{https://github.com/ctu-vras/sac-flow}.

\end{abstract}

\input{introduction}
\input{related}

\input{method2}

\input{experiments}

\input{conclusion}

\bibliographystyle{IEEEtran}
\bibliography{IEEEabrv,egbib}


\end{document}

%% file: commands.tex
\newcommand{\KITTIo}{KITTI\textsubscript{o}}
\newcommand{\KITTIv}{KITTI\textsubscript{v}}
\newcommand{\KITTIt}{KITTI\textsubscript{t}}

\newcommand\vect[1]{\ensuremath{\mathbf{#1}}}

\newcommand{\vf}{\vect{f}}
\newcommand{\bx}{\vect{x}}
\newcommand{\by}{\vect{y}}
\newcommand{\bn}{\vect{n}}
\newcommand{\cL}{\mathcal{L}}

\def\checkmark{\tikz\fill[scale=0.4](0,.35) -- (.25,0) -- (1,.7) -- (.25,.15) -- cycle;}

%% file: introduction.tex
\section{Introduction}
\begin{figure}[ht!]
    \centering
    \includegraphics[width=0.9\linewidth]{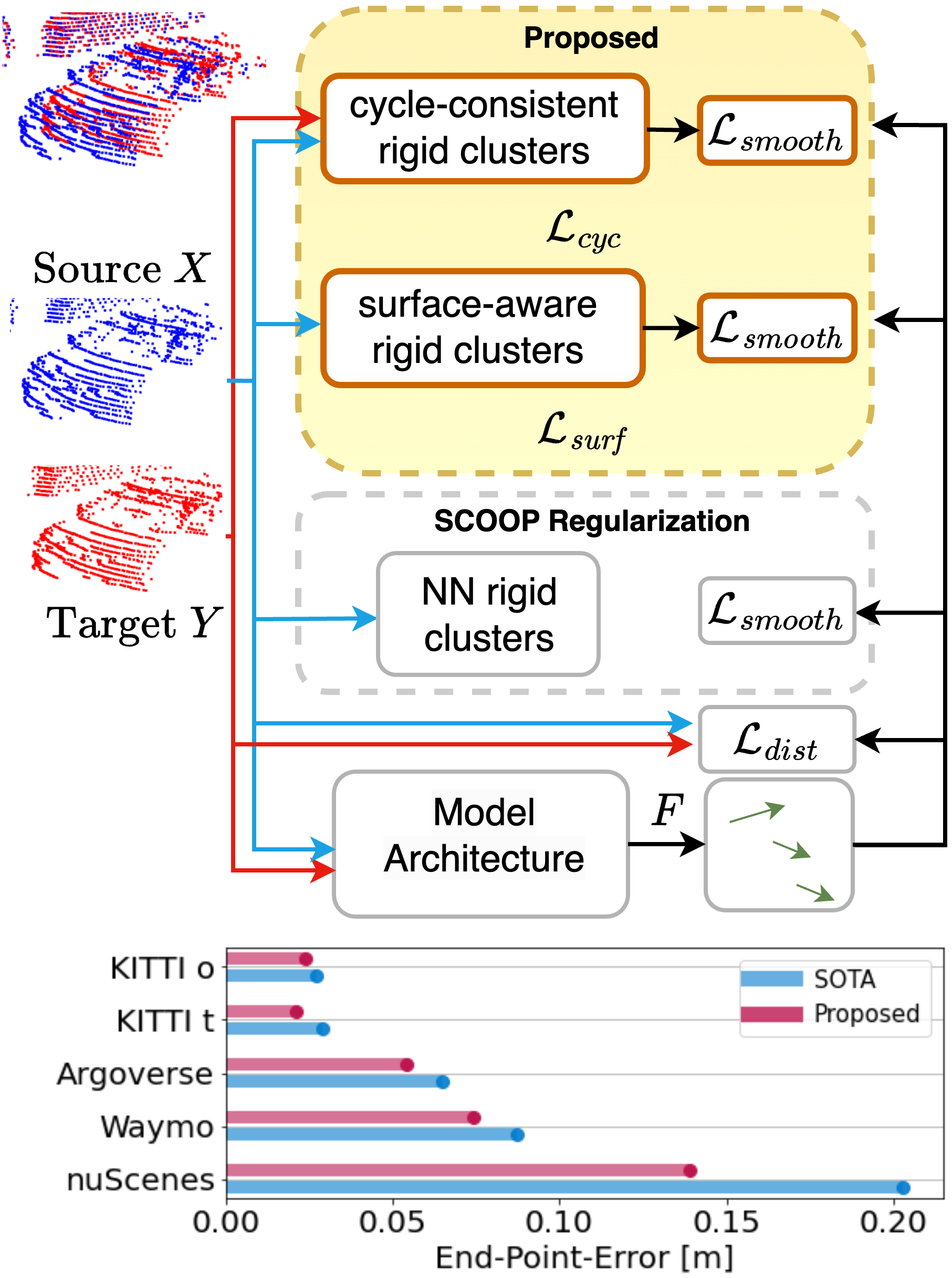}
    \caption{
    \textbf{Proposed self-supervised scene flow framework.} Self-supervised scene flow prediction is usually trained with losses that enforce the alignment of source and target point clouds and the smoothness of the flow ($\cL_{\textit{dist}}$ and $\cL_{\textit{smooth}}$ respectively). We improve the latter by introducing a surface-aware loss, $\cL_{\textit{surf}}$, and a cyclic temporal consistency one, $\cL_{\textit{cyc}}$. The proposed framework outperforms the state of the art on all tested datasets.}
    \label{fig:architercture2}
\end{figure}
%
%
%


%
\begin{figure*}[t!]
    \centering
    \includegraphics[width=0.95\linewidth]{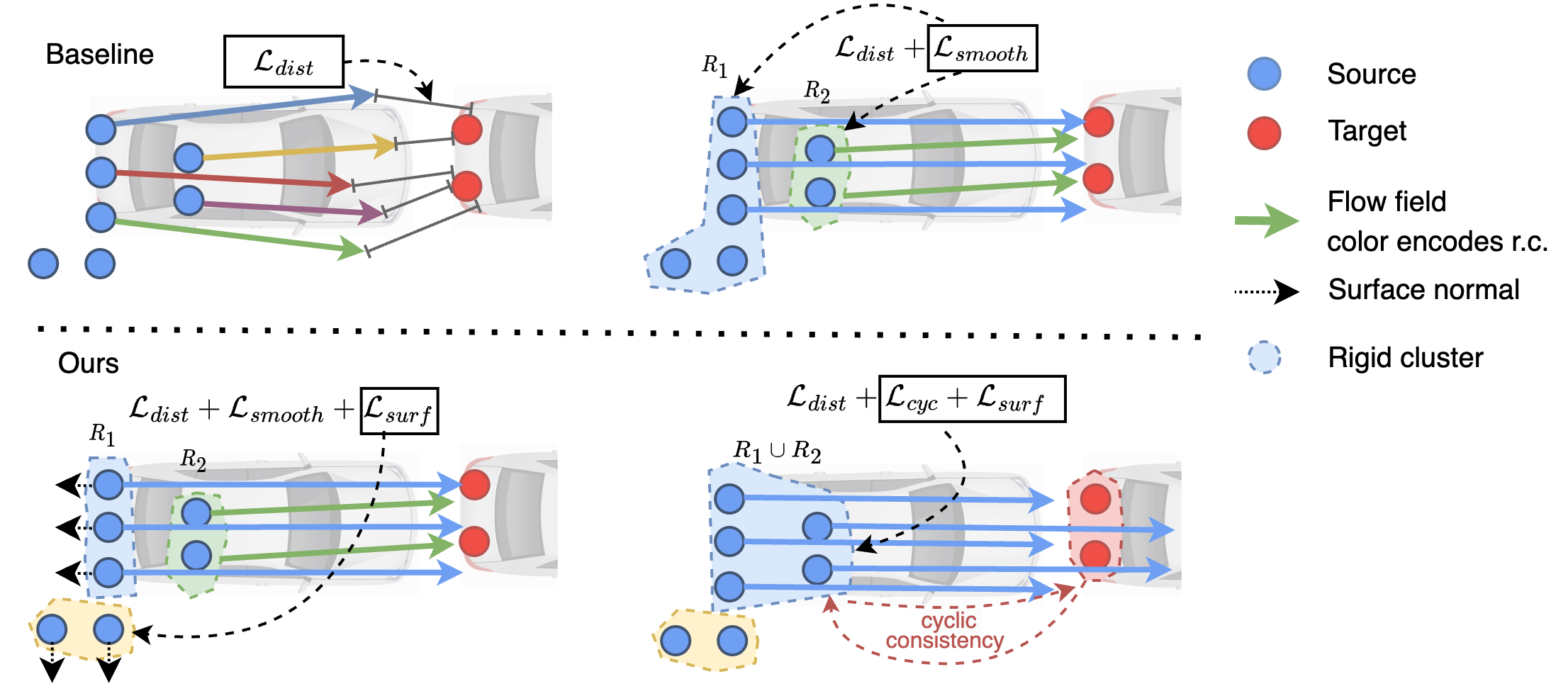}
    \caption{%
    \textbf{Illustration of baseline (top) and proposed (bottom) losses to train self-supervised scene flows}.   
    (\textit{Top-left}) Classic approaches first enforce the alignment of the two point clouds, irrespective of their structure. This results in wrong correspondences and incorrect flows.   
    (\textit{Top-right}) To improve results, local smoothness enforces motion consistency within rigid clusters. Defined only on the proximity of points, such clusters can typically be too small ($R_2$) or connect unrelated rigid bodies ($R_1$), which limits the efficiency of the smoothness loss. 
    (\textit{Bottom-left}) By taking into account surface orientation similarity in addition to spatial proximity in the definition of clusters, we mitigate the latter issue. 
    (\textit{Bottom-right}) We also propose a new cyclic consistency loss that enforces two-way time consistency between the source and target point clouds, based on significantly larger and more accurate rigid clusters. In each figure, flow vectors are colored by rigid clusters (`r.c.'), e.g., there are all colored differently when using only $\cL_{\textit{dist}}$.}
    \label{fig:overview}
\end{figure*}
%
Scene flow is defined as a three-dimensional motion field of points in the physical world. As it is a key input for many fundamental robotic and computer vision tasks, such as 
ego-motion estimation~\cite{Baur2021ICCV,tish2020}, instance segmentation~\cite{song2022ogc, DynPointCloudAnalysisEccv2022}, scene reconstruction~\cite{li2021neural} and object  detection~\cite{detflowpretrain2022},
its accurate estimation is crucial. 
%
%
Learning-based approaches to scene flow estimation were first fully supervised as in the seminal work \cite{liu:2019:flownet3d}.
%
However, ground-truth scene flows are usually synthetic, as annotations of real-world scene flows are very scarce. This makes fully-supervised methods challenging to scale and adapt to real-world use cases.
Consequently, the research community shifted its attention to self-supervised methods that learn to estimate the scene flow from unannotated sequences of point clouds, which proved to be very efficient~\cite{lang2023scoop}. 



The first mechanism at work in self-supervised methods is point cloud alignment: for each pair of consecutive point clouds in the training set, point-to-point correspondences are established based on spatio-temporal or visual similarity, and the model tries to predict flows that approximate these correspondences at best. 
If no additional constraints are considered, the predicted flows can arbitrarily deform geometric structures in the scene, which often ends up in a degenerate solution due to the high number of incorrect correspondences; see the top-left image in Figure~\ref{fig:overview} for an illustration. To avoid this,  additional regularization terms are often introduced to enforce desirable properties. The regularization term typically determines \emph{rigid clusters} (subsets of points corresponding ideally to a single rigid body) and enforces their flow to be a rigid motion via so-called \emph{smoothness}~\cite{lang2023scoop} or \emph{rigid loss}~\cite{gojcic2021weakly3dsf}; see top-right image in Figure~\ref{fig:overview}. 
However, state-of-the-art models~\cite{FlowStep3D,lang2023scoop} rely on clusters that are both over-fragmented (several clusters on the same object) and inaccurate (clusters bleeding over object borders), which limits the benefits of the regularization: one still observes unrealistic flows that deform rigid bodies substantially.

In this work, we propose to revisit how self-supervised scene flow learning is regularized, thus reaching novel state-of-the-art performance on several benchmarks. In more detail, our contributions are as follows (See Fig.\,\ref{fig:architercture2} for an overview):\\
i) We propose a novel way to form rigid clusters by explicitly considering the cyclic smoothness of the flow between source and target scans. This delivers significantly larger clusters into the smoothness loss (Figure~\ref{fig:overview}, bottom-right) \\
ii) We are first to improve the search for point correspondences in the alignment loss with the assumption of local surface regularity (Figure~\ref{fig:overview}, bottom-left). \\
iii) We compare quantitatively our framework on four publicly-available sensor-unique datasets with several baselines, including the two top-performing state-of-the-art models~\cite{li2021neural,lang2023scoop}, and we reach new state-of-the-art results in all setups.

\ifx\false
Our novel loss terms may be easily used in plug and play fashion to improve performance of other existing frameworks for flow estimation and are suitable for both learning-based and optimization methods respectively.
Our proposed methods are based on the computation of the nearest neighborhood as in other self-supervised scene flow models, ensuring similar training times.
%


\fi

%% file: related.tex
\section{Related Work}
\label{sec:related}


\paragraph{Supervised scene flow regression} During the nascent phases of learning-based 3D scene flow estimation, methodologies leaned upon synthetic datasets~\cite{MIFDB16,HPLFlowNet,Wang_2018_CVPR} for initial training. However, owing to the sim-to-real distribution gap, these approaches yield suboptimal performance when confronted with real-world point clouds~\cite{li2021neural}. FlowNet3D~\cite{liu:2019:flownet3d} adeptly incorporated principles from FlowNet~\cite{FlowNet} and underwent fully supervised training, employing L2 loss with ground-truth flow annotations. Meteor-Net~\cite{liu2019meteornet} further leveraged temporally ordered frame inputs to enhance the quality of flow inference. A continuous convolution, as introduced in~\cite{Wang_2018_CVPR}, compensated for both ego-motion and object-motion. HPLFlowNet~\cite{HPLFlowNet} turned points into a permutohedral lattice and applied bilateral convolution. FLOT~\cite{puy20flot} proposed a correspondence-based network, computing an optimal transport as an initial flow, followed by flow refinement through trained convolutions. BiPFN~\cite{BiPFN} bidirectionally propagated features from each point cloud, enriching the point feature representation. The recently introduced IHnet architecture~\cite{Wang_2023_ICCV} mitigates an iterative hierarchical network guided by high-resolution estimated information structured in a coarse-to-fine paradigm. A hierarchical network was proposed in~\cite{ding2022fast}, directly obtaining key points flow through a lightweight Trans-flow layer employing local geometry priors. 


\paragraph{Self-supervised scene flow} Recent methodologies circumvented the necessity for ground-truth flow by adopting self-supervised learning. The pioneering effort to escape full supervision emerged in \cite{Mittal_2020_CVPR}, employing a self-supervised nearest neighbor loss and ensuring cycle consistency between forward and reverse scene flow. PointPWC-Net~\cite{wu2020pointpwc} proposed a wholly self-supervised methodology, leveraging a combination of nearest-neighbors and Laplacian losses.
Flowstep3D~\cite{FlowStep3D} estimated flow with local and global feature matching using correlation matrices and iterative warping. The method called SCOOP~\cite{lang2023scoop} utilized a hybrid framework of correspondence-based feature matching, followed by a flow refinement layer using self-supervised losses.
Li et al.~\cite{Self-Point-Flow2021} confront the common challenges of neglecting surface normals and the potentiality of many-to-one correspondences in matching. They reframe the matching task as an optimal transport problem.
RigidFlow~\cite{li2022rigidflow} posits a strategy for generating pseudo scene flow within the realm of self-supervised learning. This approach hinges on piecewise rigid motion estimation applied to sets of pseudo-rigid regions, discerned through the supervoxels method~\cite{Lin2018Supervoxel}. Notably, the method emphasizes region-wise rigid alignments as opposed to point-wise alignments.
Speeding up the Neural Scene Flow stands as the focal aim in \cite{Li2023Fast}. The authors pinpoint the Chamfer distance as a computational bottleneck and reintroduce the Distance Transform as a correspondence-free loss function.
Remarkably, they attain real-time performance with precision on par with learning-based methods.
In the pursuit of self-supervised scene flow estimation, Shen et al.~\cite{Shen2023self} incorporated superpoint generation directly into the model. They implemented an iterative refinement process for both the flow and dynamically evolving superpoints.
Our method belongs to the self-supervised family, as we do not require human labels. We also regularize flow by enforcing smoothness on the pre-computed smaller groups of points compared to rigid movement on under-segmented clusters as in ~\cite{li2022rigidflow}.



\paragraph{Optimization-based flow estimation} Purely optimization-based approach Graph Prior~\cite{graphprior} utilized flow estimation without training data and Neural Prior~\cite{li2021neural} showed that flows can be optimized with structure-based prior in the form of network architecture.
The methodology presented in Scene Flow via Distillation~\cite{Vedder2023zeroflow} encapsulates a straightforward distillation framework. This approach employs a label-free optimization method to generate pseudo-labels, subsequently used for supervising a feedforward model. A new global similarity measure was introduced in \cite{Chen2022Second} in the form of a 2nd-order spatial compatibility measure; it computes a similarity between corresponding points which allows distinguishing inliers from outliers in an early stage of the point cloud registration for static scenes. 


\paragraph{Registration and rigidity regularization} Since the most standard self-supervised nearest-neighbor loss has multiple local minima, one must introduce regularization mechanisms to reach physically plausible flows, i.e., that follow well the motion of rigid structures and objects. Weak supervision in the form of ego-motion and foreground segmentation was shown to provide an object-level abstraction for the estimation of rigid flows~\cite{gojcic2021weakly3dsf}. 
Without the segmentation signals, the FlowStep3D~\cite{FlowStep3D} enforced the source point cloud to preserve its Laplacian when warped according to the predicted flow. The Laplacian was approximated by the point nearest neighborhood, forming the smoothness loss. Similarly, SLIM~\cite{Baur2021ICCV} incorporated the smoothness loss in the process of learning network weights. Meanwhile, SCOOP~\cite{lang2023scoop} integrated the smoothness loss within the optimization-based flow refinement layer. 
Implicit rigidity regularization is imposed via strong model prior in~\cite{li2021neural,pointflownet,Yi}.

We focus on meaningful regularization of self-supervised methods. To enhance the flow rigidity, we introduce novel losses that leverage the correspondence neighborhood in the target point cloud, matched to the source through the flow vectors, and introduce temporal cyclic smoothness. We also introduce surface-aware smoothness based on normals to separate close rigid clusters. These losses can be easily combined with other approaches to enhance their performance as is shown in Section~\ref{sec:exp}.



%% file: method2.tex
\section{Method}

\begin{figure*}[t]
    \centering
    \includegraphics[width=0.85\linewidth]{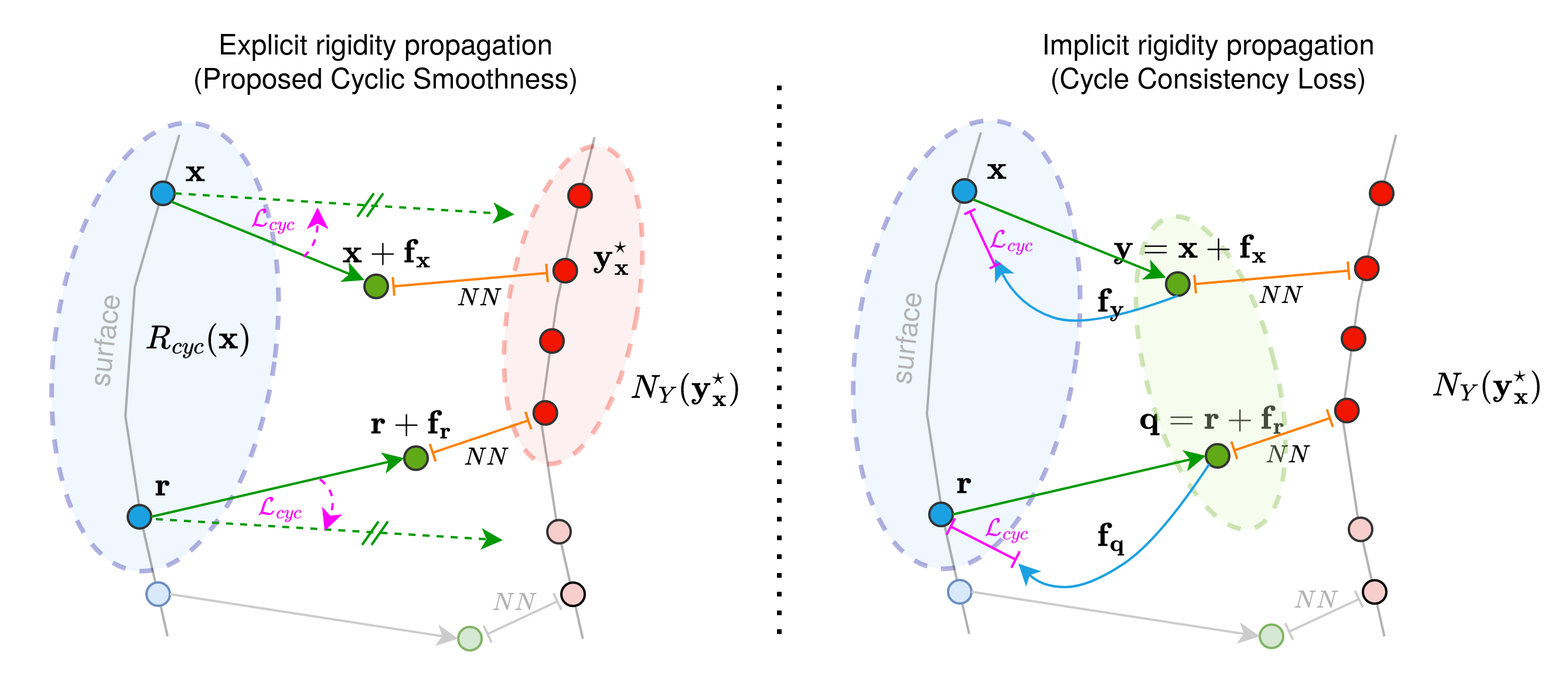}
    
    \caption{
    \textbf{Cyclic Smoothness loss.}  The new loss $\cL_{\textit{cyc}}$ enforces the same flow 
    (dashed green arrow)
    over the rigid cluster $R_{\textit{cyc}}(\bx)$ (light blue set) defined as follows: Given the source point $\bx$ and its best match $\by^\star_\bx$ in target point cloud according to flow $F$, we construct its $k$-nearest neighborhood $N_Y^k(\by_\bx^\star)$ (light red set); Any source point $\vect{r}$ whose flow $\vect{r}+\vf_{\vect{r}}$ sends it there is included in the rigid cluster $R_{\textit{cyc}}(\bx)$.
    While the proposed $\mathcal{L}_{\textit{cyc}}$ (left) \emph{explicitly} detects the rigid object as a compact cluster via normals similarity in the target point cloud and then propagates the knowledge directly to the source point cloud by enforcing rigid flow, the baseline Cycle Consistency~\cite{Mittal_2020_CVPR} $\mathcal{L}_{cycle}$ (right) \emph{implicitly} detects the rigid object by running the flow prediction in green point cloud ($\mathbf{x} + \mathbf{f_x}$) in the backward direction and then enforces the flow (blue arrows) to get back into its source.
    }
    \label{fig:cyclev2}
\end{figure*}

Given two successive point clouds $X\in \mathbb{R}^3$ and $Y\in \mathbb{R}^3$ captured from the same dynamic scene at instants $t$ and $t+\delta t$, the scene flow $F=\{\vf_\bx,\,\bx\in X\}$ at time $t$ is the set of the 3D displacements of points in \textit{source} point cloud $X$ over time interval $\delta t$, the 3D motion field in other words. The \textit{target} point cloud $Y$ is leveraged to predict this flow.      

Learning how to predict scene flows from unannotated training pairs of point clouds is classically done using an unsupervised loss that promotes point cloud alignment along with flow smoothness (under the assumption that the scene is mainly composed of rigid objects or object parts). We propose to improve such self-supervised approaches with the introduction of two novel consistency losses, as illustrated in the bottom part of Figure~\ref{fig:overview}.

\subsection{Baseline losses of self-supervised scene flow}

If no correspondence annotations are available, a typical self-supervised loss favors a flow that aligns as much as possible the source point cloud to the target one, e.g, using the nearest-neighbor distance \cite{Mittal_2020_CVPR}: 
\begin{equation}
    \cL_{\textit{dist}}(F) = \frac{1}{|X|} \sum_{\bx\in X} \min_{\by\in Y} \| \bx + \vf_\bx - \by\|^2_2. 
    \label{eq:distance}
\end{equation}
Since this loss enforces no spatial consistency whatsoever, each point is allowed to flow independently, and rigid objects can get substantially deformed or fragmented. 
To prevent this in scenes with rigid objects, one can define for each point $\bx\in X$ a \emph{rigid cluster} $R(\bx)\subset X$, i.e, a set of other points likely to lie on the same rigid fragment. A typical choice~\cite{FlowStep3D, lang2023scoop} is to simply consider the set $N^k_X(\mathbf{x})$ of $k$-nearest neighbours of $\bx$ in $X$. Once rigid clusters $R(\bx)$ are defined, the smoothness of the flow $F$ is classically enforced through a robust $L_1$ loss:
\begin{equation}
    \cL_{\textit{smooth}}(F, R) = \frac{1}{|X|} \sum_{\bx\in X} \frac{1}{|R(\bx)|} \sum_{\vect{r}\in R(\bx)}\| \vf_\bx - \vf_\vect{r}\|_1.
    \label{eq:smootness}
\end{equation}

It is desirable to have rigid clusters as large as possible to maximize the benefit of the loss, preventing rigid objects from getting deformed or fragmented. However, in the case of simply using $k$ nearest neighbors, a too-large value of $k$ incurs the risk of wrongly grouping multiple independent rigid parts into a common rigid cluster. Striking the right balance is challenging. 

Towards this end, we propose to refine the definition of the clusters so as to respect the spatio-temporal consistency of objects. In particular, we extend the definition by explicitly reconstructing the surface of objects, and by enforcing cyclic consistency of rigid clusters between source and target point clouds.

\subsection{Surface smoothness}\label{sec:L_surf}

We improve first the definition of rigid clusters by taking into account the orientation of the object's surface at each point besides its mere position. This is significant information in scenes with a prevalence of nearly planar surfaces; it will help separate two close rigid objects that have different motions. 
Denoting $\bn_\bx$ the normal of the surface captured by $X$ at position $\bx$, we define a novel surface-aware descriptor $\phi_\bx = (\bx,\bn_\bx)\in\mathbb{R}^6$ at this location. It allows us to compute a surface-aware cluster
\begin{equation}
    R_{\textit{surf}}(\vect{x}) = N^k_\Phi(\phi_\vect{x}),
\end{equation}
where $\Phi= \{\phi_\bx,\,\bx\in X\}$. In practice, we find $k_n$ nearest neighbors of the point $\vect{x}$, followed by obtaining principal vectors of covariance matrices of each of the points in the neighborhood. The main principal vector normalized to one corresponds to the normal vectors $\mathbf{n_x}$, which is used to construct surface-aware descriptor $\phi$. Grouping similar surfaces like this improves local rigidity. Then, we follow the method described in~\cite{Tombari10} for sign disambiguation. The implementation is available in the open-source PyTorch3D library~\cite{pytorch3d}. 
The new surface-aware smoothness loss then reads:
\begin{equation}
    \cL_{\textit{surf}}(F) = \cL_{\textit{smooth}}(F, R_{\textit{surf}}).
    \label{eq:smootness-normals}
\end{equation}
By using this loss instead of the regular one, flow consistency is enforced among pairs of scene elements that are close and similarly oriented.

\subsection{Cyclic smoothness}\label{sec:L_csm}
We now introduce a second smoothing loss that enforces cyclic (forward-backward) consistency between times $t$ and $t+\delta t$. It aims at propagating the information about object rigidity from the target point cloud back to the source one via forward correspondences, see Figure \ref{fig:cyclev2}. 

Following matching loss (\ref{eq:distance}), a given scene flow $F$ effectively matches each source point $\bx$ with a point $\by^\star\in Y$:
\begin{equation}
    \by^\star = \arg\min_{\by\in Y}\|\bx + \vf_\bx - \by\|_2^2.
\end{equation}
Based on this correspondence, we define the \textit{cyclic rigid cluster} of point $\vect{x}$ as a set of points $\vect{r}\in X$ whose displaced position according to $F$ falls within the neighborhood of $\vect{y}^\star_\vect{x}$. More formally:
\begin{equation}
    R_{\textit{cyc}}(\bx)= \{\vect{r}\,|\, \vect{r}+\vf_\vect{r} \in  N^k_Y(\by^\star_\bx)\}.
\end{equation}
The corresponding cycle-consistent loss reads: 
\begin{equation}
    \cL_{\textit{cyc}}(F) = \cL_{\textit{smooth}}(F, R_{\textit{cyc}}).
    \label{eq:cycle-smootness}
\end{equation}
Note that the definition of this new loss does not introduce any extra parameters. 

\subsection{Training objective and Architecture}

Given an architecture that can be trained to predict a scene flow $F$ from a pair $(X,Y)$ of point clouds, we can make use of the novel proposed losses.  In that case, the training objective becomes: 
\begin{equation}
    \cL = \mathcal{L}_{\textit{dist}} + \alpha_{\textit{surf}} \cL_{\textit{surf}} + \alpha_{\textit{cyc}} \cL_{\textit{cyc}},
\end{equation}
where $\alpha_{\textit{surf}}$ and $\alpha_{\textit{cyc}}$ are hyperparameters balancing the contribution of individual loss terms. 
This approach is architecture-independent and can thus be combined with existing models. In practice, we put it at work with two existing scene flow models, namely SCOOP~\cite{lang2023scoop} stereo-based datasets and Neural Prior~\cite{li2021neural} for real LiDARs.

\color{black}

%% file: experiments.tex
\begin{table*}[t]
  \centering
  \caption{\textbf{Comparative results of scene flow estimation methods on stereo-based datasets.} 
    We evaluate scene flow based on standard metrics \textit{EPE}, \textit{AS}, \textit{AR} and \textit{Out}, for different settings of supervision, train data and test data. Combined with two architectures (SCOOP and Neural Prior), our approach beats all fully-supervised baselines trained on FT3D as well as all self-supervised methods. 
      There are differences between the reported performance (`*') of SCOOP and the one recomputed from original codebase (`\dag'), which surpasses the former. To be fair in evaluation, we report both performances while optimizing the flow until convergence to achieve the models' upper-bound performance. 
      Our proposed loss terms improve the performance even further. For our method, metrics are averaged over 6 runs.}
  %
  \scalebox{0.95}{
    \begin{tabular}{lccclcccc}
    \toprule
    Method & Supervision &  Train data &  Test data & \textit{EPE}\,[m] $\downarrow$ & \textit{AS}\,[\%] $\uparrow$ & \textit{AR}\,[\%] $\uparrow$ & \textit{Out.}\,[\%] $\downarrow$ \\
    \midrule
    FlowNet3D\cite{FlowNet} & Full & FT3D & \KITTIo & 0.173 & 27.6 & 60.9 & 64.9 \\
    FLOT\cite{puy20flot}  & Full & FT3D & \KITTIo & 0.107 & 45.1 & 74.0 & 46.3 \\
    BiPFN\cite{BiPFN} & Full & FT3D & \KITTIo & 0.065 & 76.9 & 90.6 & 26.4 \\
    R3DSF\cite{gojcic2021weakly3dsf} & Full & FT3D & \KITTIo & 0.042 & 84.9 & 95.9 & 20.8 \\
    \midrule
    FlowStep3D\cite{FlowStep3D} & Self & FT3D & \KITTIo & 0.102 & 70.8 & 83.9 & 24.6 \\
    SCOOP*\cite{lang2023scoop} & Self & FT3D & \KITTIo & 0.047 & \underline{91.3} & \underline{95.0} & 18.6 \\
    SCOOP\dag\cite{lang2023scoop} & Self & FT3D & \KITTIo & \underline{0.037} & 89.4 &  94.9 & \underline{18.0} \\
    \textbf{SCOOP + Ours} & Self & FT3D & \KITTIo & \textbf{0.024} \textcolor{LimeGreen}{(-35\%)} & \textbf{97.1} & \textbf{98.6} & \textbf{13.9} \\
    %
    \midrule
    \midrule
    Graph Prior\cite{graphprior} & Self & N/A & \KITTIt & 0.082 & 84.0 & 88.5 & - \\
    Neural Prior\cite{li2021neural}  & Self & N/A & \KITTIt & \underline{0.036} & \underline{92.3} & \underline{96.2} & \underline{13.2} \\
    \textbf{Neural Prior + Ours} & Self & N/A & \KITTIt & \textbf{0.034} \textcolor{LimeGreen}{(-5\%)} & \textbf{92.5} & \textbf{97.5} & \textbf{12.9}  \\

    \midrule
    JGWTF\cite{Mittal_2020_CVPR} & Self & nuScenes + \KITTIv & \KITTIt & 0.105 & 46.5 & 79.4 & - \\ 
    SPF\cite{Shen2023self}  & Self & KITTI\textsubscript{r} + \KITTIv & \KITTIt & 0.089 & 41.7 & 75.0 & - \\
    SCOOP*\cite{lang2023scoop} & Self & KITTI\textsubscript{v} & \KITTIt &  0.039 & 93.6 & 96.5 & 15.2 \\
    SCOOP\dag\cite{lang2023scoop} & Self & KITTI\textsubscript{v} & \KITTIt & \underline{0.029} & \underline{95.9} & \underline{98.0} & \underline{12.2} \\
    \textbf{SCOOP + Ours} & Self & KITTI\textsubscript{v} & \KITTIt & \textbf{0.021} \textcolor{LimeGreen}{(-28\%)} & \textbf{98.9} & \textbf{99.5} & \textbf{11.3} \\ 
    
    \bottomrule
    \end{tabular}%
    }
  \label{tab:stereo}%
\end{table*}%

\section{Experiments}
\label{sec:exp}

In this section, we demonstrate quantitatively and qualitatively the performance gains stemming from our proposed losses on two state-of-the-art architectures and four benchmarks. 
We also analyze these gains, as well as our design and parameter choices, in ablation studies.




\paragraph{Datasets and metrics.}
We perform experiments on four scene flow datasets. The first one is stereoKITTI~\cite{kittisf}, which contains real-world self-driving scenes. We used the commonly benchmarked subset released by \cite{liu:2019:flownet3d} consisting of point clouds created from stereo images. The dataset, dubbed KITTI\textsubscript{o}, was split by \cite{Mittal_2020_CVPR} into train and test parts denoted KITTI\textsubscript{v} and KITTI\textsubscript{t} respectively. Additionally, the authors introduced an unlabelled dataset, KITTI\textsubscript{r}, consisting of actual LiDAR sensor data from the same driving sequences. 
Next, we used three large-scale LiDAR autonomous driving datasets, namely Argoverse~\cite{Argoverse}, nuScenes~\cite{nuscenes}, and Waymo~\cite{Sun_2020_CVPR}, with challenging dynamic scenes captured by different LiDAR sensors. The LiDAR datasets were sampled and processed according to \cite{li2021neural} for the fair comparison. Since there are no official scene flow annotations, we followed the data processing method in \cite{ScalableWaymo2022} to collect pseudo-ground-truth scene flows from object detection annotations as done in \cite{li2021neural}. Ground points were removed for every dataset as in \cite{liu:2019:flownet3d}. We use \textbf{full} range of the LiDAR points in input and evaluation.
%
%


\begin{figure}[t]
    \centering
    \includegraphics[width=\linewidth]{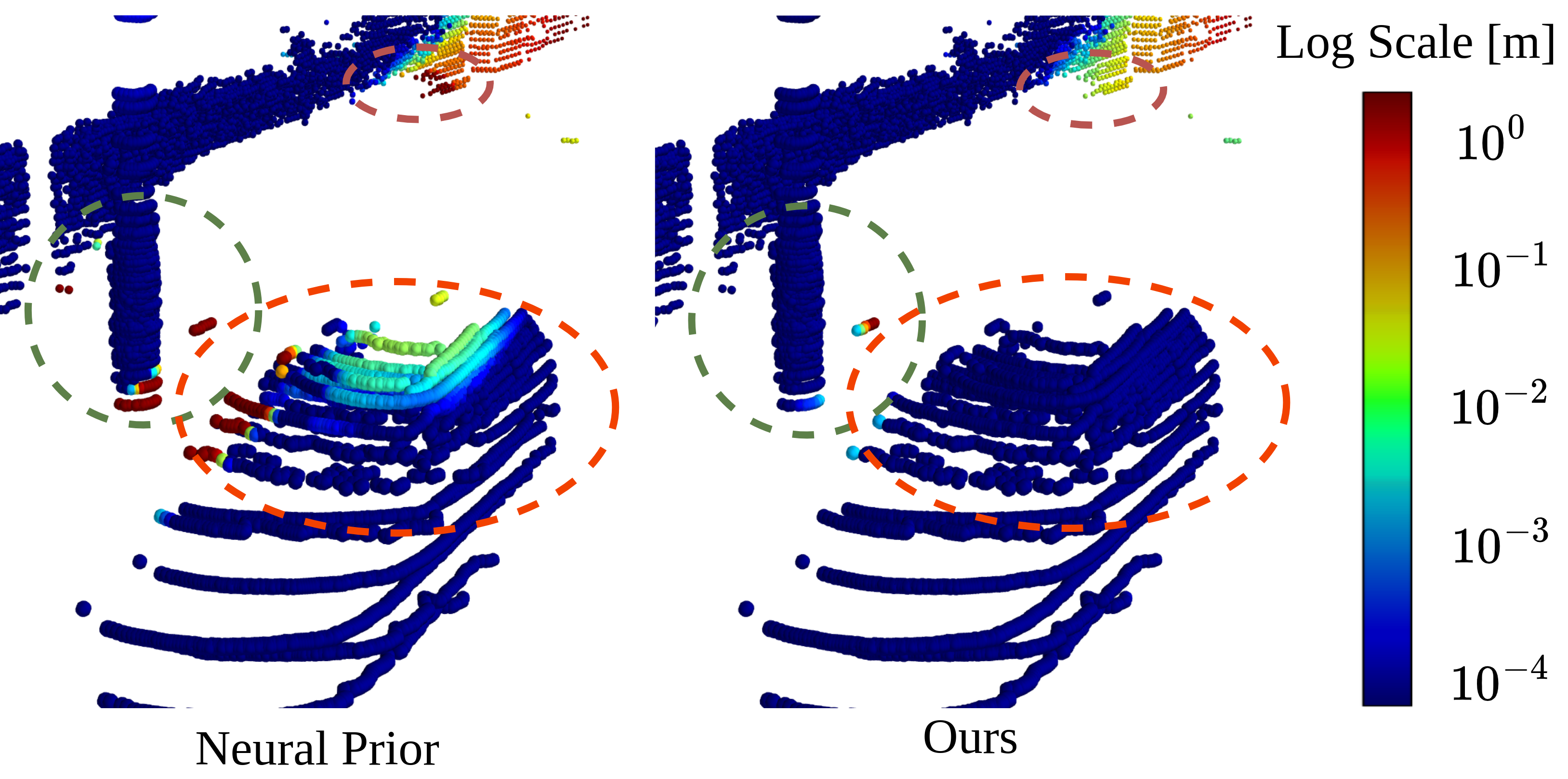}
    \caption{\textbf{Example of improvements brought by proposed framework on real LiDAR data}.  In this scene from the Argoverse dataset, we show the per-point flow estimation error encoded by color on a logarithmic scale. While the Neural Prior~\cite{li2021neural} baseline on the left fails to produce consistent flows along the majority of the vehicle body (orange ellipse), the addition of our proposed losses corrects the full body rigidity. The same applies to the pole (green ellipse) and the wall in the back (red ellipse).}
    \label{fig:argo-qual}
\end{figure}

We use well-established evaluation metrics \cite{li2021neural,lang2023scoop,puy20flot} for scene flow. Namely End-Point-Error (\textit{EPE}), Strict Accuracy (\textit{AS}), Relaxed Accuracy (\textit{AR}), Angle Error ($\theta$) and Outliers (\textit{Out.}). These metrics are based on the point error $e_\bx$ and the relative error $e^{\text{rel}}_\bx$:
\[
e_\bx = \|\vf_\bx - \vf^{\text{gt}}_\bx\|_2 , 
\quad e^{\text{rel}}_\bx = \frac{{\|\vf_\bx - \vf^{\text{gt}}_\bx\|}_2}{\|\vf^{\text{gt}}_\bx\|_2},
\]
where $\vf_\bx$ and $\vf^{\text{gt}}_\bx$ are the predicted and ground-truth motions for point $\bx$ in the source point cloud. 
The \textit{EPE} is the average point error in meters and is considered to be the main metric for scene flow evaluation; Strict accuracy \textit{AS} is the percentage of points which either have an error
$e_\bx < 0.05$m or relative error $e^{\text{rel}}_\bx < 5\%$; Similarly, the \textit{AR} is the percentage of points for which the error satisfies either $e_\bx < 0.1$m or $e^{\text{rel}}_\bx < 10\%$; 
The outlier metric (\textit{Out.}) is the percentage of points with error either $e_\bx > 0.3$m or $e^{\text{rel}}_\bx > 10\%$;
Finally, $\theta$ is the mean angular error between the estimated and ground-truth scene flows.
\begin{table}[]
\setlength{\tabcolsep}{3pt}
    \centering
    \caption{\textbf{Performance on LiDAR datasets.} We show performance on standard LiDAR benchmarks with commonly reported metrics. Fully-supervised methods are trained on FT3D\textsubscript{o} dataset. When adding the proposed losses, we consistently improve the performance of the state-of-the-art Neural Prior method across all datasets. Our method results are averages over 6 runs.}
    \begin{tabular}{l|c|lrrc}
    \toprule
         Method & D & \textit{EPE}\,[m]$\downarrow$ & \textit{AS}\,[\%]$\uparrow$ & \textit{AR}\,[\%]$\uparrow$  
          & $\theta$\,[rad]$\downarrow$ 
         \\
    \midrule
    FlowNet3D\cite{FlowNet}  & \multirow{5}{*}{\rotatebox[origin=c]{90}{Argoverse}} & 0.455 & 1.34 & 6.12 & 0.736 \\
    JGWTF\cite{Mittal_2020_CVPR}  & & 0.542 & 8.80 & 20.28 & 0.715 \\
    PointPWC-Net\cite{wu2020pointpwc}  &  & 0.409 & 9.79 & 29.31 & 0.643 \\
    Neural Prior\cite{li2021neural}  &  & \underline{0.065} & \underline{77.89} & \underline{90.68} & \underline{0.230} \\
    \textbf{Neural Prior + Ours}  &   & \textbf{0.054} \textcolor{LimeGreen}{(-17\%)} & \textbf{81.11} & \textbf{92.51} & \textbf{0.223} \\
    \midrule
    FlowNet3D\cite{FlowNet}  &  \multirow{5}{*}{\rotatebox[origin=c]{90}{Nuscenes}}& 0.505 & 2.12 & 10.81 & 0.620 \\
    JGWTF\cite{Mittal_2020_CVPR}  & & 0.625 & 6.09 & 13.90 & 0.432\\
    PointPWC-Net\cite{wu2020pointpwc}  &  & 0.431 & 6.87 & 22.42 & 0.406 \\
    Neural Prior\cite{li2021neural}  & & \underline{0.203} & \underline{49.64} & \underline{76.03} & \underline{0.244}\\
    \textbf{Neural Prior + Ours}  & & \textbf{0.139} \textcolor{LimeGreen}{(-32\%)} & \textbf{55.56} & \textbf{80.43} & \textbf{0.220} \\
    \midrule
    R3DSF\cite{gojcic2021weakly3dsf}  & \multirow{3}{*}{\rotatebox[origin=c]{90}{Waymo}}& 0.414 & 35.47 & 44.96 & 0.527 \\
    Neural Prior\cite{li2021neural}  & & \underline{0.087} & \underline{78.96} & \underline{89.96} & \underline{0.300}\\
    \textbf{Neural Prior + Ours}  &  & \textbf{0.074} \textcolor{LimeGreen}{(-15\%)} & \textbf{81.65} & \textbf{91.45}  & \textbf{0.290} \\
    \bottomrule
    \end{tabular}
    \label{tab:lidar}
\end{table}
\paragraph{Baselines, settings, implementation details.}
We benchmark our method with other 3D scene flow frameworks such as fully-supervised methods trained on synthetic FT3D dataset~\cite{FT3D} with ground-truth flows~\cite{liu:2019:flownet3d, FlowStep3D,puy20flot}, weakly-supervised models with access to the foreground segmentation labels~\cite{gojcic2021weakly3dsf}, self-supervised models~\cite{lang2023scoop,Mittal_2020_CVPR,wu2020pointpwc}, and optimization-based approaches~\cite{li2021neural,graphprior}.
%

%
Point clouds in both FT3D and stereoKITTI are obtained similarly by lifting stereo images to 3D through ground-truth 2D optical flow. Hence, their domain gap is relatively small, and they offer cross-dataset evaluation, enabling models to be trained on synthetic FT3D in a fully-supervised fashion.
On the other hand, in LiDAR-based autonomous driving scenarios, point clouds are much sparser and provide different sampling patterns with no direct one-to-one correspondences, resulting in a much more challenging setting for scene flow estimation. 
To be consistent with the evaluation of top-performing scene flow models, we test our method on all points present in the scene and not partial chunks and optimize until the convergence of the loss or until reaching the maximum predefined number of iterations.
%
We incorporate in our framework the two top-performing architectures SCOOP~\cite{lang2023scoop} and Neural Prior~\cite{li2021neural}, which are the current state-of-the-art respectively on stereoKITTI and on LiDAR scene flow datasets.


%
Our method is implemented in PyTorch and trained on one NVIDIA Tesla A100 GPU. We use the default parameters for both architectures, as reported in the original papers. We use $\alpha_{\textit{surf}}=10$ as SCOOP does for the regular weight of smoothness on the stereoKITTI dataset, and we similarly set $\alpha_{\textit{cyc}} = 10$. For Neural Prior, we use $\alpha_{\textit{surf}}=1$ since there is already a regularization included in the network structure, and we keep $\alpha_{\textit{cyc}}=10$ for consistency in all experiments. 

For the stereoKITTI dataset, we use $k=32$ for $k$-nn computations as in the original SCOOP configuration and $k=4$ for LiDAR datasets since LiDAR point clouds are much more sparse. Lastly, we compute normals for surface smoothness based on $k_n=5$ nearest neighbors for all datasets.


\begin{table}[t]
    \centering
    \caption{\textbf{Influence of proposed losses.} The performance reached with various loss combinations is evaluated on nuScenes dataset with Neural Prior architecture. We observe that, by gradually adding the proposed losses, we improve the performance in all metrics.}
    \scalebox{0.9}{
    \begin{tabular}{cccccc}
         \toprule
         $\cL_{\textit{smooth}}$ & $\cL_{\textit{cyc}}$ & $\cL_{\textit{surf}}$ & \textit{EPE}\,[m] $\downarrow$ & \textit{AS}\,[\%] $\uparrow$
         & \textit{AR}\,[\%] $\uparrow$ \\ 
         \midrule
         & & & 0.203 & 49.70 & 76.03 \\ 
         \checkmark & & & 0.175 &  53.22 & 79.11  \\
         & \checkmark & & 0.163 & 50.04 & 76.47 \\
         & & \checkmark & 0.166 & 55.18 & 80.39  \\
         \checkmark & \checkmark & & 0.145 & 53.59 & 79.27  \\
         
          & \checkmark & \checkmark & \textbf{0.139} & \textbf{55.56} & \textbf{80.43} \\

         \bottomrule
    \end{tabular}
    }
    \label{tab:ablation-loss}
\end{table}

\subsection{Evaluation on StereoKITTI}

In Table \ref{tab:stereo}, we show the results of our method with tested architectures on the stereoKITTI dataset in comparison to baselines. Coupled with existing self-supervised architectures, our losses outperform the fully-supervised models trained on large synthetic FT3D datasets and the self-supervised and optimization-based methods. Mainly, our self-supervised loss terms improve the state-of-the-art architectures on both \KITTIo{} and \KITTIt{} splits. On these two splits, our proposed loss terms deliver relative \textit{EPE} gains of $35.1\%$ and $27.5\%$, respectively, on top of state-of-the-art SCOOP architecture. Performance is boosted in all other metrics as well. These results demonstrate the merit of the proposed losses: they bring current state-of-the-art scene flow models to new performance levels.



\subsection{Generalization over various LiDAR datasets}
In Table \ref{tab:lidar}, we demonstrate the benefit of our loss terms applied in combination with Neural Prior architecture on different LiDAR datasets. While we keep the proposed parameters unchanged for all the datasets in the paper, we achieve consistent improvement in all cases with various sensor configurations, even on much sparser nuScenes LiDAR. We improved the performance of Neural Prior architecture in all of the evaluation metrics. 
Such consistent gains demonstrate that the assumptions of surface rigidity and regularity that underpin our losses are agnostic to the sensor domain and sampling pattern. See Figure \ref{fig:argo-qual} for a qualitative sample.

\subsection{Ablations studies}

\begin{table}[]
    \centering
    \resizebox{!}{1cm}{
    \begin{tabular}{c|c|c|c}
         Regularization & \textit{EPE}[m] & \textit{AS}[\%] & \textit{AR}[\%] \\
         \toprule
         $\textcolor{blue}{\mathcal{L}_{cycle}}$ & 0.168 & 49.36 & 76.56 \\
         $\textcolor{red}{\mathcal{L}_{\textit{cyc}}}$ & 0.163 & 50.04 & 76.47 \\

         $\textcolor{blue}{\mathcal{L}_{cycle}}$ + $\textcolor{blue}{\mathcal{L}_{sm}}$ &  0.149 & 53.08 & 79.30\\
            
         $\textcolor{red}{\mathcal{L}_{\textit{cyc}}}$ + $\textcolor{blue}{\mathcal{L}_{sm}}$ & 0.145 & 53.59 & 79.57 \\
         $\textcolor{blue}{\mathcal{L}_{cycle}}$ + $\textcolor{red}{\mathcal{L}_{\textit{cyc}}}$ + $\textcolor{blue}{\mathcal{L}_{sm}}$ & 0.132 & 55.41 & 80.76\\
          
         \bottomrule
    \end{tabular}
    }
    \caption{Result of cycle losses on nuScenes, without $\mathcal{L}_{\textit{surf}}$ to isolate the effect of $\mathcal{L}_{cyc}$. Standard losses in \textcolor{blue}{blue} and our proposed component in \textcolor{red}{red}.}
    \label{tab:cycle-regularization}
\end{table}

\begin{figure}[t]
    \centering
    \includegraphics[width=0.99\linewidth]{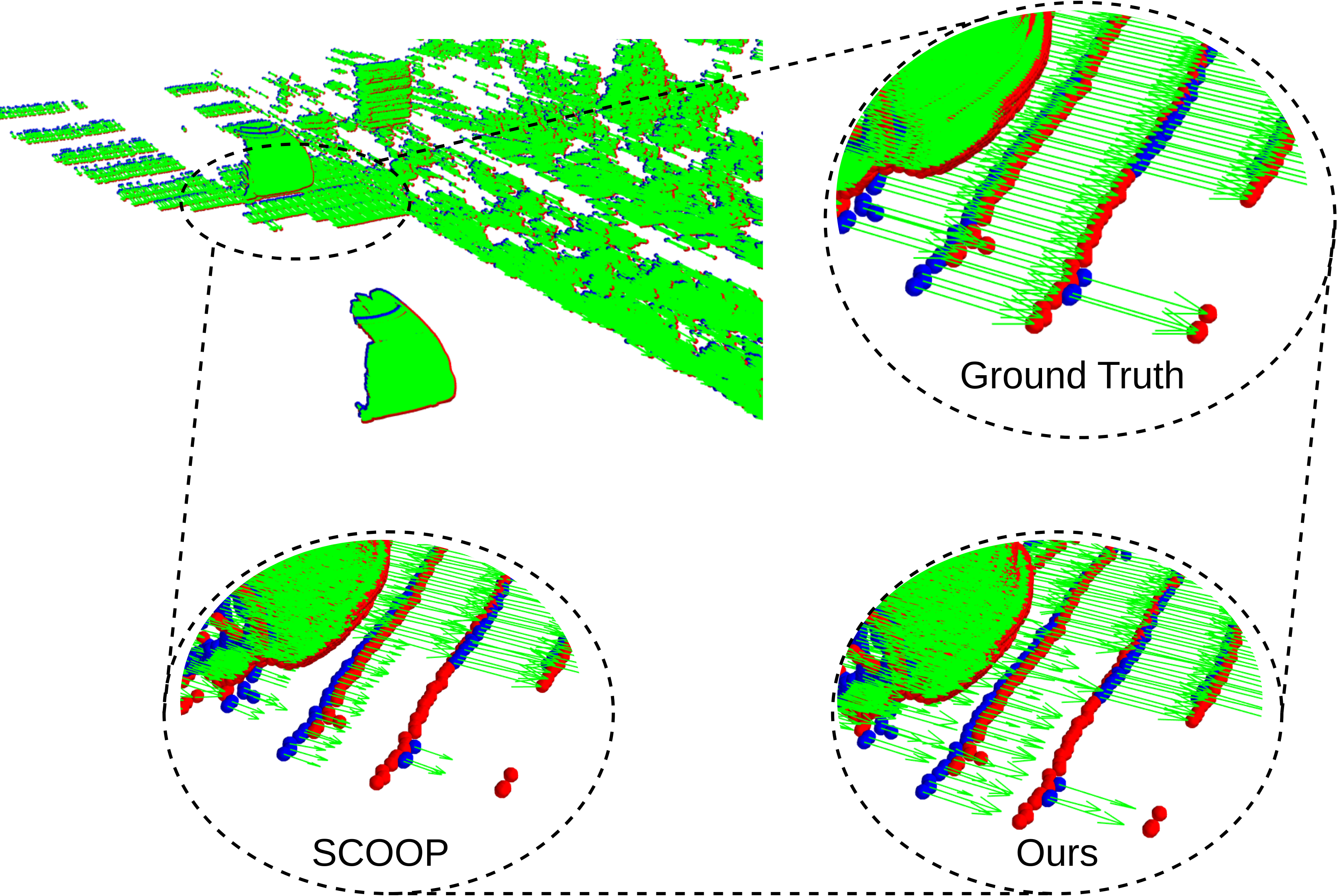}
    \caption{\textbf{Qualitative example on \KITTIt.} Top-left viewpoint shows a point cloud with flow after ground filtering with some ground falsely segmented. As a result, the ground with flow induced by ego movement shares the same neighborhood in smoothness loss. Our surface loss, however, groups points, taking into account the associated normals, thus separating the planar ground from the vehicle for a much more consistent flow.}
    \label{fig:kittit-qual}
\end{figure}

\paragraph{Influence of loss terms.}
We examine the performance of various combinations for proposed losses with Neural Prior architecture on the nuScenes dataset; see Table \ref{tab:ablation-loss}. We observe the benefit of adding the smoothness loss to the existing prior regularization. Our proposed terms $\cL_{\textit{cyc}}$ and $\cL_{\textit{surf}}$ increase the performance even further when adding components separately, where $\cL_{\textit{cyc}}$ has a large impact on \textit{EPE} and $\cL_{\textit{surf}}$ greatly increases \textit{AS}.
Note that $\cL_{\textit{surf}}$ substitutes the $\cL_{\textit{smooth}}$, and therefore combination of both at the same time is omitted.
\paragraph{Cyclic Smoothness vs. Cycle Consistency}
The name of our proposed Cyclic Smoothness implies similarities with Cycle Consistency loss introduced in JGWTF~\cite{Mittal_2020_CVPR}. We demonstrate differences in Figure~\ref{fig:cyclev2} and experimentally verify gains in Table~\ref{tab:cycle-regularization}.
Both losses regularize the flow predicted in the blue source point cloud by transferring the knowledge about the object rigidity observable in the target point cloud, see Figure~\ref{fig:cyclev2}. The main difference between proposed cyclic smoothness $\mathcal{L}_{\textit{cyc}}$ and standard cycle consistency $\mathcal{L}_{cycle}$
stems from the way the rigid object in the red point cloud is discovered.  Both approaches can prevent trivial failure cases in which the flow of two distinct blue points ends up in a single red point; however, in more complicated cases, the behavior is different.
The implicit rigidity propagation is extremely dependent on the presence of a high-density point cloud and the small noise of the predicted (green) point cloud in order to discover the actual rigid object successfully.
We experimentally demonstrate that both proposed losses are complementary; see Table~\ref{tab:cycle-regularization}.
%
%
The best results are achieved with the combination of both, while the second best results are achieved with the proposed explicit propagation. 
%
%
\paragraph{Influence of Surface Smoothness on stereoKITTI}
\begin{figure}[t]
    \centering
    \includegraphics[width=0.95\linewidth]{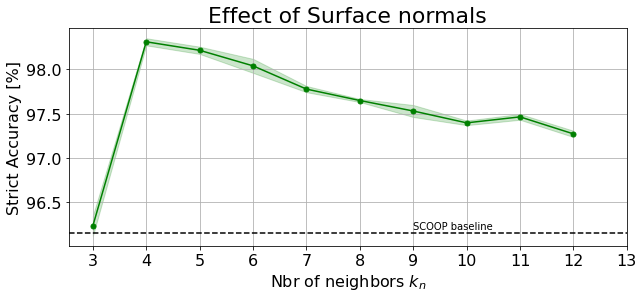}
    \caption{\textbf{Ablation of normal estimation on \KITTIt.} Influence on the performance of the number of neighboring points used to compute surface normals. The best performance for the SCOOP model is obtained with $k_n=4$, and further increase of neighborhood diminishes the performance.}
    \label{fig:ablation-plot-normals}
\end{figure}
We ablate the number $k_n$ of nearest neighbor points needed to construct normals for $\cL_{\textit{surf}}$ on \KITTIt~dataset in Figure \ref{fig:ablation-plot-normals}. We observe that the best performance is reached on the stereo-based dataset when four neighbors are used to compute normals. Adding more points diminishes the performance yet still surpasses SCOOP; see Figure \ref{fig:kittit-qual} for an example of separating flows of vehicle and falsely segmented ground. The performance is exceptionally worse when we utilize the minimum number of points for normal estimation, i.e. \textit{three}. We attribute this to the sampling bias of stereo camera data, where the three points most likely lie on a line that comes from adjacent pixels, making normal estimation ill-posed.

%



         

\section{Limitations}
By increasing the number of points in rigid clusters in $\cL_{\textit{cyc}}$, we implicitly enforce the flow to be a purely translational movement rather than a more realistic rotation and translation. However, due to the high framerate of the sensor, the translation movement is a sufficient approximation for most of the applications. Note also that this issue is not specific to our approach as it is already a weak point of the standard smoothness loss. 

If the point cloud is very sparse, $\mathcal{L}_{\textit{surf}}$ would only introduce noise to the features for neighborhood calculation in final smoothness. Flow estimation also would not benefit from $\mathcal{L}_{\textit{surf}}$ on planar objects.

%% file: conclusion.tex
\section{Conclusion}
Self-supervised learning of scene flow without ground-truth signals is prone to degenerative solutions. We have introduced surface awareness and cyclic smoothness into the self-supervised learning framework, which proved essential for scene flow regularization and improved the performance of state-of-the-art models. The experiments demonstrated that the method works in real-world LiDAR settings, with point clouds from a standard stereo-based dataset, and with multiple network architectures, proving its generalization ability.
Since our method aims for an improved selection of the object-centric rigid clusters, the proposed loss terms could also be adapted to instance segmentation in the future. We also plan to extend the cycle smoothness to work across multiple frames while seeking a broader temporal consistency.